\documentclass[twoside]{article}

\usepackage[accepted]{aistats2025}
\usepackage{tikz}
\usepackage{standalone}
\usepackage{bm}
\usepackage{amsmath}
\usepackage{amssymb}
\usepackage{nicefrac}
\usepackage[inline]{enumitem}
\usepackage{siunitx}
\usepackage{booktabs}
\PassOptionsToPackage{hyphens}{url}\usepackage[colorlinks=true, linkcolor=red, citecolor=blue, urlcolor=blue]{hyperref}
\usepackage[capitalise]{cleveref}

\DeclareMathOperator*{\argmax}{arg\,max}
\DeclareMathOperator*{\argmin}{arg\,min}

\tikzstyle{pdeblock} = [rectangle, rounded corners, minimum width=3cm, minimum height=1cm,text centered, draw=black, fill=TUred!30]
\tikzstyle{process} = [rectangle, minimum width=3cm, minimum height=1cm, text centered, draw=black, fill=TUorange!30]
\tikzstyle{randvar} = [rectangle, minimum width=3cm, minimum height=1cm, text centered, draw=black, fill=TUdarkblue!30]
\tikzstyle{gnvar} = [rectangle, minimum width=1cm, minimum height=1cm, text centered, draw=black, fill=TUgray!30]
\tikzstyle{arrow} = [thick,->,>=stealth]

\usepackage[
  obeyFinal,
  textsize=tiny,
  tickmarkheight=0.1cm,
  figwidth=0.99\linewidth,
]{todonotes}
\definecolor{TUred}{RGB}{165,30,55}
\definecolor{TUgold}{RGB}{180,160,105}
\definecolor{TUdark}{RGB}{50,65,75}
\definecolor{TUgray}{RGB}{175,179,183}

\definecolor{TUdarkblue}{RGB}{65,90,140}
\definecolor{TUblue}{RGB}{0,105,170}
\definecolor{TUlightblue}{RGB}{80,170,200}
\definecolor{TUlightgreen}{RGB}{130,185,160} 
\definecolor{TUgreen}{RGB}{125,165,75}
\definecolor{TUdarkgreen}{RGB}{50,110,30}
\definecolor{TUocre}{RGB}{200,80,60}
\definecolor{TUviolet}{RGB}{175,110,150}
\definecolor{TUmauve}{RGB}{180,160,150}
\definecolor{TUbeige}{RGB}{215,180,105}
\definecolor{TUorange}{RGB}{210,150,0}
\definecolor{TUbrown}{RGB}{145,105,70}

\def\R{\mathbb{R}}
\def\N{\mathbb{N}}

\usepackage[round]{natbib}

\begin{document}

\runningtitle{Flexible and Efficient Probabilistic PDE Solvers through GMRFs}

\twocolumn[

\aistatstitle{Flexible and Efficient Probabilistic PDE Solvers \\ through Gaussian Markov Random Fields}

\aistatsauthor{ Tim Weiland \And Marvin Pförtner \And Philipp Hennig }

\aistatsaddress{ University of Tübingen, Tübingen AI Center } ]

\begin{abstract}
  Mechanistic knowledge about the physical world is virtually always expressed via partial differential equations (PDEs). %
  Recently, there has been a surge of interest in probabilistic PDE solvers---Bayesian statistical models mostly based on Gaussian process (GP) priors which seamlessly combine empirical measurements and mechanistic knowledge.
  As such, they quantify uncertainties arising from e.g.~noisy or missing data, unknown PDE parameters or discretization error by design.
  Prior work has established connections to classical PDE solvers and provided solid theoretical guarantees.
  However, scaling such methods to large-scale problems remains a challenge, primarily due to dense covariance matrices.
  Our approach addresses the scalability issues by leveraging the Markov property of many commonly used GP priors.
  It has been shown that such priors are solutions to stochastic PDEs (SPDEs), which, when discretized, allow for highly efficient GP regression through sparse linear algebra.
  In this work, we show how to leverage this prior class to make probabilistic PDE solvers practical, even for large-scale nonlinear PDEs, through greatly accelerated inference mechanisms.
  Additionally, our approach also allows for flexible and physically meaningful priors beyond what can be modeled with covariance functions.
  Experiments confirm substantial speedups and accelerated convergence of our physics-informed priors in nonlinear settings.
\end{abstract}

\section{INTRODUCTION}
\begin{figure*}[t]
  \centering
  {\color{TUdark}
  \begin{tikzpicture}[node distance=2cm]
    \filldraw[fill=TUdarkblue!10!white, draw=TUdark,dashed, rounded corners] (1.8,0.9) rectangle(14.5, -3.5);
    \node[anchor=east] at (14, -3.1) {\textcolor{TUdark!100}{\textbf{Prior construction}}};
    \node[draw, align=center] (pde) [pdeblock] {PDE \\ $\frac{\partial u}{\partial t} + u \frac{\partial u}{\partial x} = \nu \frac{\partial^2 u}{\partial x^2}$};
    \node[draw, align=center] (lin-proxy) [process, right of=pde, xshift=2cm] {Linearized Proxy \\ $\frac{\partial u}{\partial t} + c \frac{\partial u}{\partial x} = \nu \frac{\partial^2 u}{\partial x^2}$};
    \draw [arrow] (pde) -- (lin-proxy);
    \node[draw, align=center] (spde) [process, right of=lin-proxy, xshift=2cm] {SPDE \\ $\frac{\partial u}{\partial t} + c \frac{\partial u}{\partial x} - \nu \frac{\partial^2 u}{\partial x^2} = \tau \mathcal{W}$};
    \node[draw, align=center] (noise) [process, right of=spde, xshift=2cm] {Noise \\ $\mathcal{W}$};
    \draw [arrow] (lin-proxy) -- (spde);
    \draw [arrow] (noise) -- (spde);
    \node[draw, align=center] (spatial-disc) [process, below of=lin-proxy, yshift=0.4cm] {Spatial discretization \\ FEM};
    \node[draw, align=center] (temporal-disc) [process, below of=noise, yshift=0.4cm] {Temporal discretization \\ Implicit Euler};
    \node[draw, align=center] (gmrf-prior) [randvar, below of=spde, yshift=-0.25cm] {GMRF Prior \\ $u \sim \mathcal{N}(\bm{0}, \bm{Q}^{-1})$};
    \draw [arrow] (spde) -- (gmrf-prior);
    \draw [arrow] (spatial-disc) -- (gmrf-prior);
    \draw [arrow] (temporal-disc) -- (gmrf-prior);

    \filldraw[fill=TUdarkgreen!10!white, draw=TUdark,dashed, rounded corners]
    (-2, -2.0) -- (1.8, -2.0) -- (1.8, -3.5) -- (14.5, -3.5) -- (14.5, -8.1) -- (-2, -8.1) -- cycle;
    \node[anchor=east] at (14, -7.7) {\textcolor{TUdark!100}{\textbf{Inference}}};
    \node[draw, align=center] (data-gmrf) [randvar, below of=gmrf-prior, yshift=-1.0cm] {$u_{\text{Data}} \sim \mathcal{N}(\bm{\mu_{u \mid y}}, \bm{Q_{u \mid y}}^{-1})$};
    \draw [arrow] (gmrf-prior) -- (data-gmrf);
    \node[fill=TUdarkgreen!10!white, inner sep=1pt] at (8, -3.9) {mechanistic knowledge};
    \node[draw, align=center] (data) [process, right of=data-gmrf, xshift=2.5cm] {Direct measurements $y$ \\ e.g.\ Initial condition};
    \node[anchor=south] at (data.north) {empirical knowledge};
    \draw [arrow] (data) -- (data-gmrf);
    \node[draw, align=center] (pde-disc) [process, below of=pde, yshift=-1.0cm] {Discretization \\ e.g.\ Collocation};
    \filldraw[fill=TUviolet!40!white, draw=TUdark,rounded corners] (-1.75,-4.0) rectangle(3.75, -6.5);
    \node[anchor=south west] at (0.2, -6.25) {\scriptsize $k=1,2,\dots$};
    \node[draw, align=center] (gn) [gnvar, below of=pde-disc, yshift=-0.25cm] {Gauss-Newton \\ Step};
    \node[draw, align=center] (muk) [gnvar, right of=gn, xshift=1.0cm] {$\bm{\mu_k}$};
    \draw [arrow] (data-gmrf) -- node [above] {$\bm{\mu_0} = \bm{\mu_{u \mid y}}$} (muk);
    \draw [arrow] (pde) -- (pde-disc);
    \draw [arrow] (pde-disc) -- (gn);
    \draw [arrow] (muk.north) |- ++(0.0, 0.5) -| (gn.north);
    \draw [arrow] (gn.south) |- ++(0.0,-0.5) -| (muk.south);
    \node[draw, align=center] (mustar) [gnvar, below of=gn, yshift=-0.1cm, xshift=1.25cm] {$\bm{\mu_*}$};
    \node[draw, align=center] (xfinal) [randvar, below of=data-gmrf, yshift=-0.1cm] {$u_{\text{Data+PDE}} \sim \mathcal{N}(\bm{\mu_{*}}, \bm{Q_{\text{LA}}}^{-1})$};
    \draw [arrow] (gn.south) |- node [left] {converged} (mustar);
    \draw [arrow] (mustar) -- node [below] {Gauss-Newton Laplace} (xfinal);
    \node[anchor=south] at (xfinal.north) {Posterior};
    \node[anchor=north,xshift=2mm,fill=TUdarkgreen!10!white,inner sep=1pt,yshift=-0.7mm] at (pde-disc.south) {computational knowledge};
  \end{tikzpicture}}
  \caption{A flowchart of the high-level steps we propose for prior construction and inference. This conceptual sketch uses Burgers' equation as a concrete example for a simple nonlinear PDE.}
  \label{fig:flowchart}
\end{figure*}
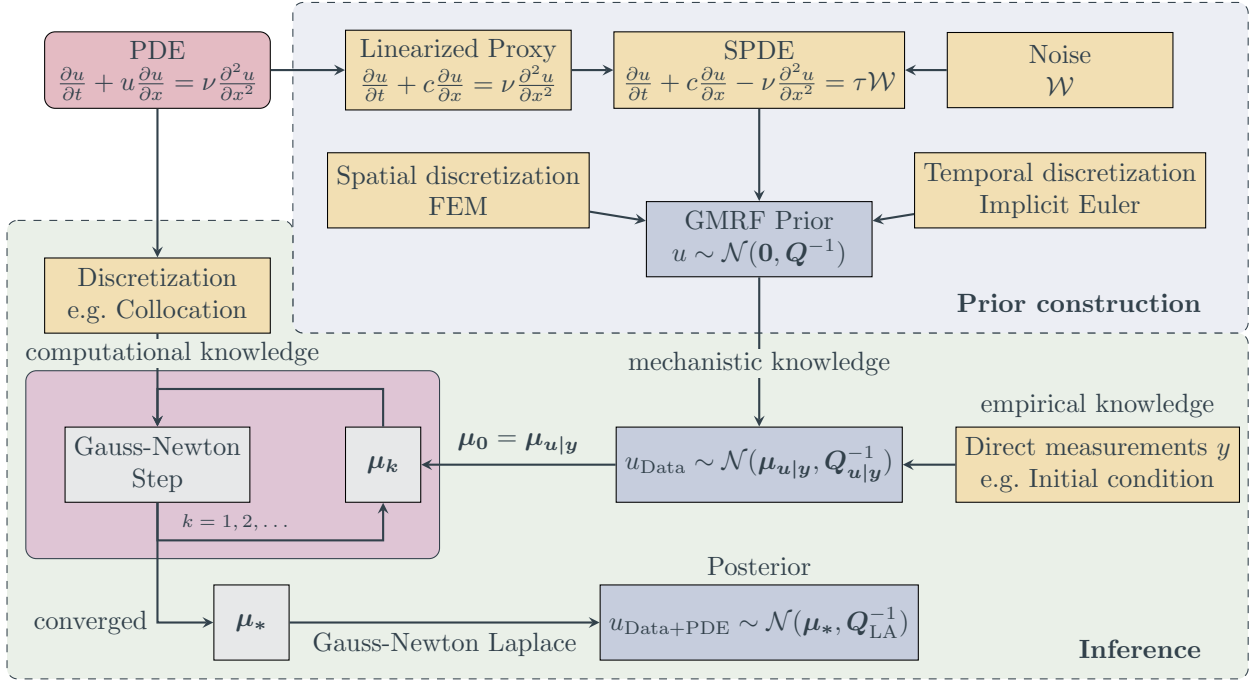
In the physical sciences, the laws of nature are mostly expressed through partial differential equations (PDEs).
A PDE describes how a function, modelling a quantity of interest, changes over space and time through its partial derivatives.
Due to their expressive power, PDEs have found applications in many scientific domains, e.g.\ climate and weather forecasting \citep{Bauer2015}, oceanography \citep{Chandrasekar2022}, continuum mechanics \citep{Reddy2013} and electromagnetics \citep{Solin2005}.
As closed-form solutions are typically unavailable, applications employ numerical PDE solvers which compute approximate solutions.

Both in the machine learning and scientific domain communities, there has been an increasing interest in physics-informed machine learning methods \citep{Karniadakis2021} which learn not only from direct measurements, but also from mechanistic knowledge codified in PDEs.
A common promise of such methods is that they predict accurate solutions to PDEs at a fraction of the cost of classic solvers.
However, recent worrying reproducibility problems with deep-learning based simulation methods show that a careful connection between classic PDE solvers and machine learning methodology is urgently needed \citep{McGreivy2024}. %

Nevertheless, neither classical nor most machine-learned simulation methods handle \textbf{uncertainty} gracefully.
In real-world applications, simulations of physical systems are subject to various types of uncertainties:
\begin{enumerate*}[label=(\roman*)]
  \item Sensors produce noisy, sparse, and potentially incomplete measurements of the system state.
  \item The parameters of the PDE are hardly ever known exactly.
  \item The solution is only an approximated and typically discretized.
\end{enumerate*}

Probabilistic PDE solvers are probabilistic numerical methods \citep{hennig2022probabilistic} that treat such uncertainties fundamentally by interpreting the approximation of a PDE solution as a Bayesian inference problem.
The most prominent approach is based on Gaussian process priors that are conditioned to solve a discretization of the PDE.
Under specific design choices, these methods exactly reproduce classic numerical solvers in the posterior mean \citep{Pfoertner2022}, thus inheriting the convergence guarantees of these methods.

This provides a strong theoretical argument for probabilistic PDE solvers over less structured machine-learned surrogates. However, high computational cost has so far restricted the practical value of general probabilistic solvers. The underlying reason for this is that previous, especially theoretically motivated works, strive for formal generality of the probabilistic framework, to maximize the quantification of discretization error, parameter uncertainty, and noise in the data. Lowering the computational cost of probabilistic PDE solvers to be more in line with that of classic solvers requires a trade-off between flexible and slow, or restrictive and fast models and algorithms.
In an example of this direction, some prior works on scalable probabilistic solvers are based on restrictive prior assumptions, e.g.\ tensor product covariance functions \citep{Weiland2024GPFVM, Wang2021}. These lower cost, but severely limit the expressive power of the prior.

\textbf{Our central contribution is to equip probabilistic PDE solvers with faster inference mechanisms under flexible, physics-informed priors.}
The foundation of our method is an approach widely used in spatial statistics: Compact basis function discretizations of stochastic PDEs (SPDEs) yield finite-dimensional approximations of stochastic processes with sparse precision matrices, enabling substantial computational gains \citep{Lindgren2011}.
Our proposed method consists of two stages, which are summarized in \cref{fig:flowchart}.
First, we construct a prior: In \cref{sec:prior-construction}, we demonstrate how existing approaches based on Gaussian process priors inspire finite-dimensional approximations.
This will then naturally lead to the idea of using physics-informed correlation models as a strong alternative to widely used stationary models.
Having constructed a prior, we then revisit in \cref{sec:inference-mechanisms} the general notion of probabilistic numerics, to phrase Bayesian inference on PDE discretizations as merely yet another source of data, analogous to statistical measurements, and show how to concretely realise it in our specific setting.
As we will outline, this viewpoint enables highly efficient inference mechanisms for spatiotemporal, nonlinear PDEs.
Finally, we validate our method empirically in \cref{sec:experiments}.

\section{PRIOR CONSTRUCTION}
\label{sec:prior-construction}
\subsection{GP-based PDE solvers}
As a starting point, we briefly outline the approach that is currently used to design probabilistic PDE solvers.
In spatiotemporal settings, an initial boundary value problem (IBVP) consists of a PDE along with initial and boundary conditions.
Consider as an example an IBVP based on Burgers' equation:
\begin{align}
\frac{\partial u}{\partial t} + u \frac{\partial u}{\partial x} &= \nu \frac{\partial^2 u}{\partial x^2}, \label{eq:burgers-pde} \\
  u(0, x) &= \phi(x), \label{eq:burgers-ic} \\
  u(t, 0) &= u(t, 1), \label{eq:burgers-periodic}
\end{align}
where \cref{eq:burgers-ic} with $\phi: [0, 1] \rightarrow \R$ constitutes the initial condition and \cref{eq:burgers-periodic} the (periodic) boundary condition.

Probabilistic PDE solvers frame the problem of finding the solution to \crefrange{eq:burgers-pde}{eq:burgers-periodic} as a Bayesian inference problem by positing a Gaussian process (GP) prior $u \sim \mathcal{GP}(\mu, k)$ over the unknown solution function.
The prior mean $\mu$ and covariance function $k$ model prior knowledge about the solution.
Observe that discretizing the initial condition \cref{eq:burgers-ic} simply yields direct observations of the solution, which may be integrated with classic GP regression.
This yields a posterior $u^{*} \sim \mathcal{GP}(\mu^{*}, k^{*})$ with covariance
\begin{multline}
  k^{*}(\bm{z_1}, \bm{z_2}) = k(\bm{z_1}, \bm{z_2}) \\ - k(\bm{z_1}, \bm{X}) \left( k(\bm{X}, \bm{X}) + \sigma^2 \bm{I} \right)^{-1} k(\bm{X}, \bm{z_2}), \label{eq:gp-posterior-cov}
\end{multline}
with discretization points $\bm{X} \in \R^{N \times 2} (N \in \N)$ and measurement noise $\sigma^2 > 0$.

In fact, GPs are not only closed under point observations, but---under mild assumptions---also under other linear transformations \citep{Pfoertner2022}.
The resulting posterior covariance has the same structure as \cref{eq:gp-posterior-cov}, but in the downdate term occurences of $\bm{X}$ are replaced with applications of the linear operator \citep[see][Section 4]{Pfoertner2022}.
Differentiation is a linear operation, and thus, if the PDE of interest is linear, the posterior under a PDE discretization is available in closed form.
In the nonlinear case, discretizations may be formulated as nonlinear transformations of linear operations.
The posterior is then generally no longer a GP, but it may be approximated, e.g.~through a Laplace approximation \citep{Chen2024}.

\textbf{Practical considerations.}
These properties make GPs theoretically appealing for obtaining approximate PDE solutions under statistical data.
In practice however, we encounter two issues:
\begin{enumerate}
    \item \textbf{The expression of prior knowledge} is restricted to what can be modelled tractably with covariance functions. A standard choice is the Mat\'ern covariance function.
    \item \textbf{Scalability} of these models is impeded significantly by the (typically dense) matrix inversion on the right-hand side of \cref{eq:gp-posterior-cov}. Approaches to circumvent these issues exist, but may require even stronger prior assumptions such as separable covariance functions.
\end{enumerate}

\textbf{Gauss-Markov priors and SDEs.}
In the context of probabilistic ODE solvers, the solution to these problems is  the use of Markovian priors which are solutions to linear SDEs \citep{Tronarp2019}.
This enabled the use of filtering and smoothing algorithms for linear-time inference.
Unfortunately, a direct transfer of these ideas to spatiotemporal processes is not possible.
However, in the following we leverage the same underlying ideas:
We consider multivariate Markov priors which are solutions to linear S\textbf{P}DEs and demonstrate the benefits of this viewpoint for probabilistic PDE solvers.

\subsection{SPDE Priors}
The SPDE approach, as introduced by \citet{Lindgren2011}, is motivated by the example of the Mat\'ern process.
It is based on the insight that a Gaussian process with Mat\'ern covariance function is the stationary solution of the Whittle-Mat\'ern SPDE \citep{Whittle1954}
\begin{equation}
  \label{eq:whittle-matern-spde}
  \underbrace{\left( \kappa^2 - \Delta \right)^{\nicefrac{\alpha}{2}}}_{=: \mathcal{D}} u(\bm{x}) = \mathcal{W}(\bm{x})
\end{equation}
on $\R^d$, where $\mathcal{W}$ is a Gaussian white noise process, $\kappa > 0, \nu > 0, \alpha = \nu + \nicefrac{d}{2}$ and $\Delta$ is the Laplacian operator.

In practice, the solution of the SPDE is approximated by a parametric Gaussian process $u(\bm{x}) \approx \sum_{i=j}^N u_j \phi_j(\bm{x})$ with compactly supported basis / feature functions $\phi_j$ and $\bm{u} \sim \mathcal{N}(\bm{0}, \bm{Q}^{-1})$.
To solve for the corresponding precision matrix $\bm{Q}$, the SPDE is typically discretized using the finite element method (FEM), which results in a linear system
\begin{equation}
  \bm{K} \bm{u} = \bm{w}
\end{equation}
with stochastic right-hand side $\bm{w} \sim \mathcal{N}(\bm{0}, \bm{M})$.
The system matrix $\bm{K} \in \R^{N \times N}$ is given by the stiffness matrix corresponding to $\mathcal{D}$ and $\{\phi_j\}_{j = 1}^N$ and $\bm{M} \in \R^{N \times N}$ is the mass matrix corresponding to the basis functions.
Since the basis functions $\phi_j$ have compact support, both $\bm{K}$ and $\bm{M}$ are sparse.
The solution of the linear system is hence given by
\begin{equation}
  \bm{u} \sim \mathcal{N}(\bm{0}, \underbrace{\bm{K}^{-1} \bm{M} \bm{K^{-T}}}_{= \bm{Q}^{-1}}).
\end{equation}
In FEM, it is common to approximate the mass matrix $\bm{M}$ by a diagonal matrix $\bm{\tilde{M}}$ through mass lumping.
This results in a sparse precision matrix
\begin{equation}
  \bm{Q} \approx \bm{K}^T \bm{\tilde{M}}^{-1} \bm{K}. \label{eq:sparse-prec-matrix}
\end{equation}
Due to the sparsity of its precision matrix, $\bm{u}$ is called a Gaussian Markov Random Field (GMRF).
GMRF inference is quite efficient, which has lead to widespread use in spatial statistics \citep{Lindgren2022}.

\subsection{Physics-Informed Priors}
\label{sec:physics-informed-priors}

\begin{figure}
  \centering
  \begin{tikzpicture}
    \filldraw[fill=TUred!10!white, draw=none] (-0.5, 0.5) rectangle (7.5, 4.2);
    \node[anchor=south east] at (7.5, 0.5) {Advection-diffusion SPDE};
    \filldraw[fill=TUorange!10!white, draw=none] (-0.5, -0.5) rectangle (7.5, -4.2);
    \node[anchor=north east] at (7.5, -0.5) {product Mat\'ern};

    \node[anchor=center] at (1.85, 0) {$t=\SI{1.5}{sec}$};
    \node[anchor=center] at (5.5, 0) {$t=\SI{3.0}{sec}$};

    \node[shape=rectangle, draw=TUdark, minimum width=2cm, rounded corners, fill=white] (prior) at (0, 0) {\includegraphics[width=1.5cm]{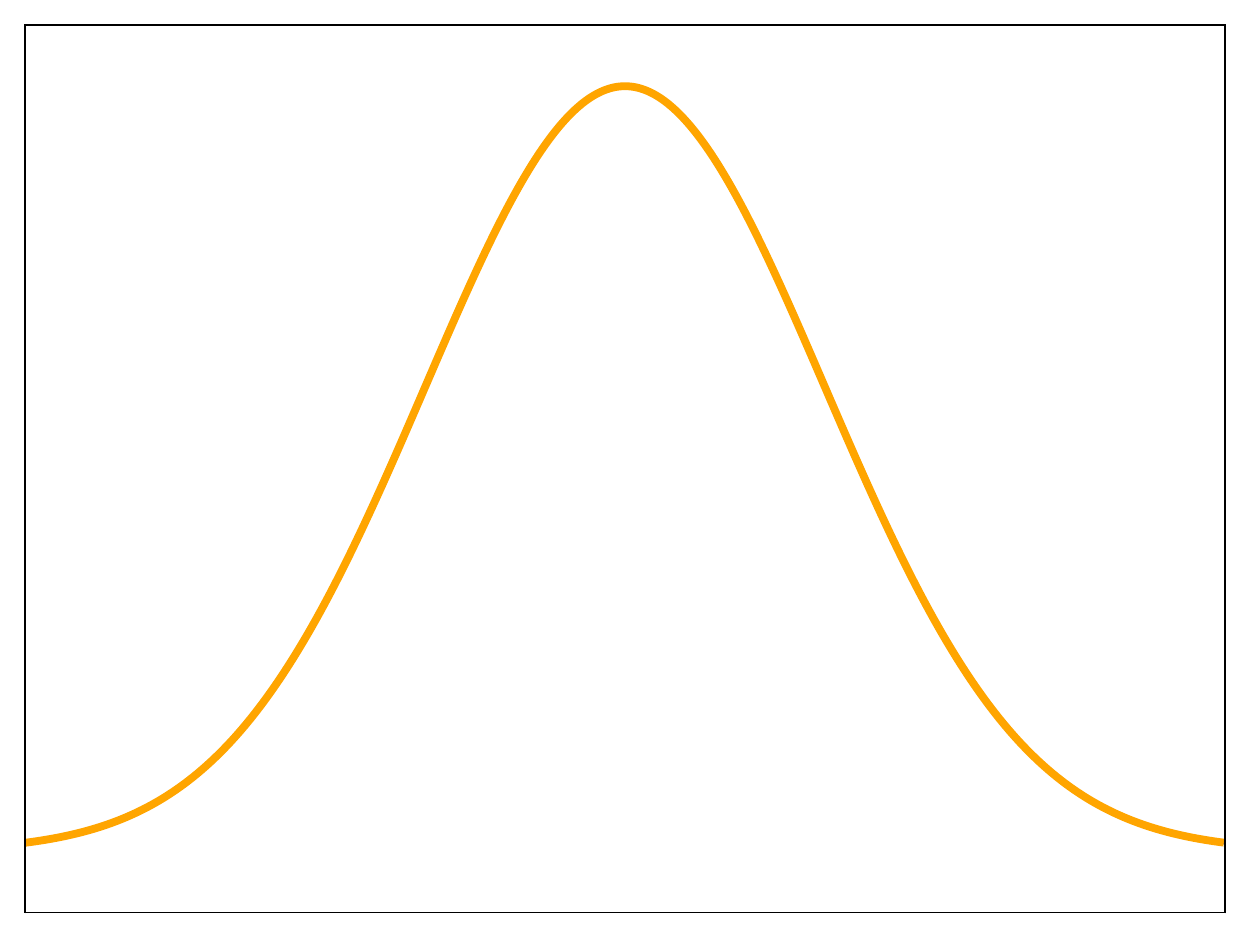}};

    \node[shape=rectangle, draw=TUdark, minimum width=2cm, rounded corners] (advdiff1) at (1.5, 2.5) {\includegraphics[width=3.5cm]{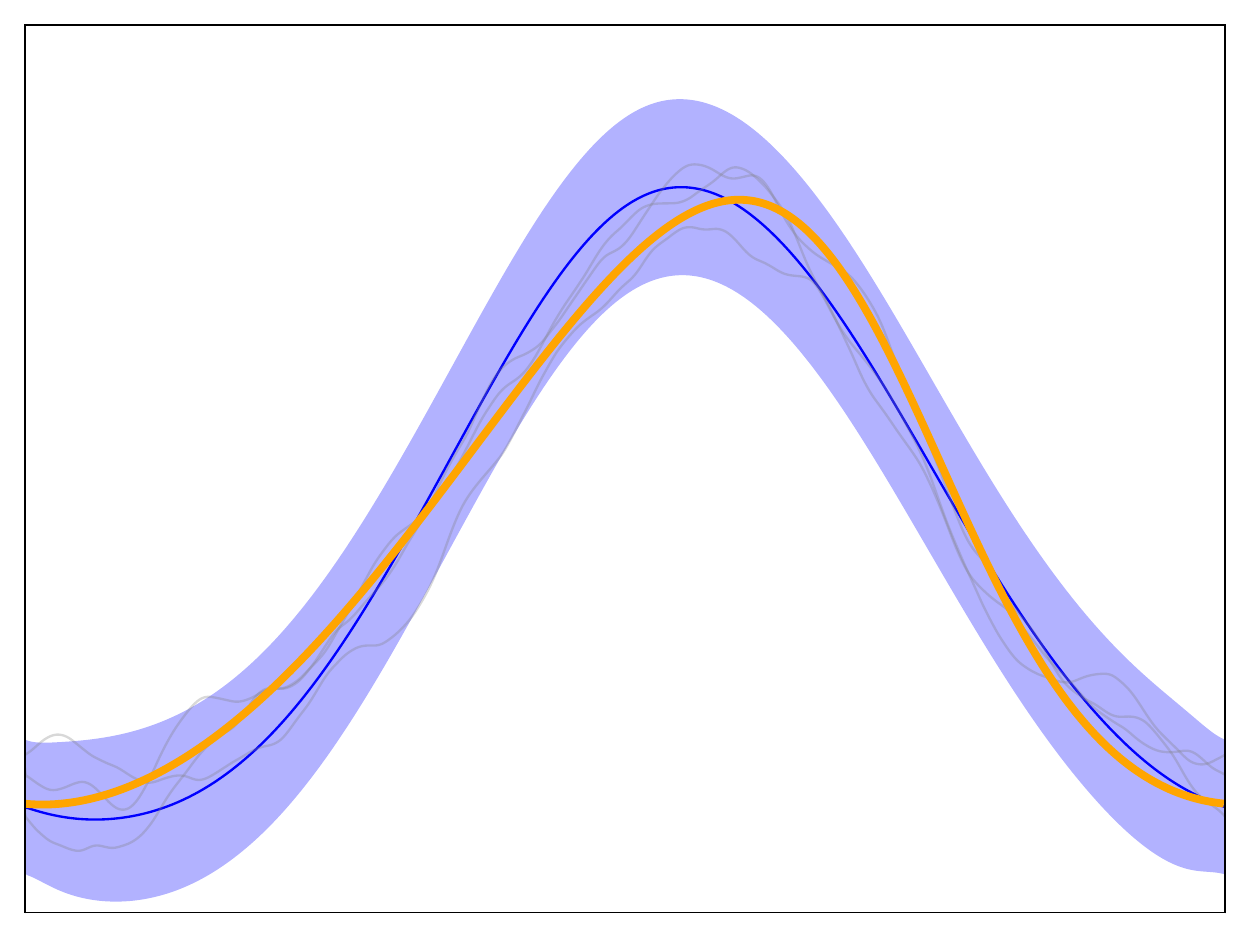}} edge[<-] (prior);
    \node[shape=rectangle, draw=TUdark, minimum width=2cm, rounded corners] (advdiff2) at (5.5, 2.5) {\includegraphics[width=3.5cm]{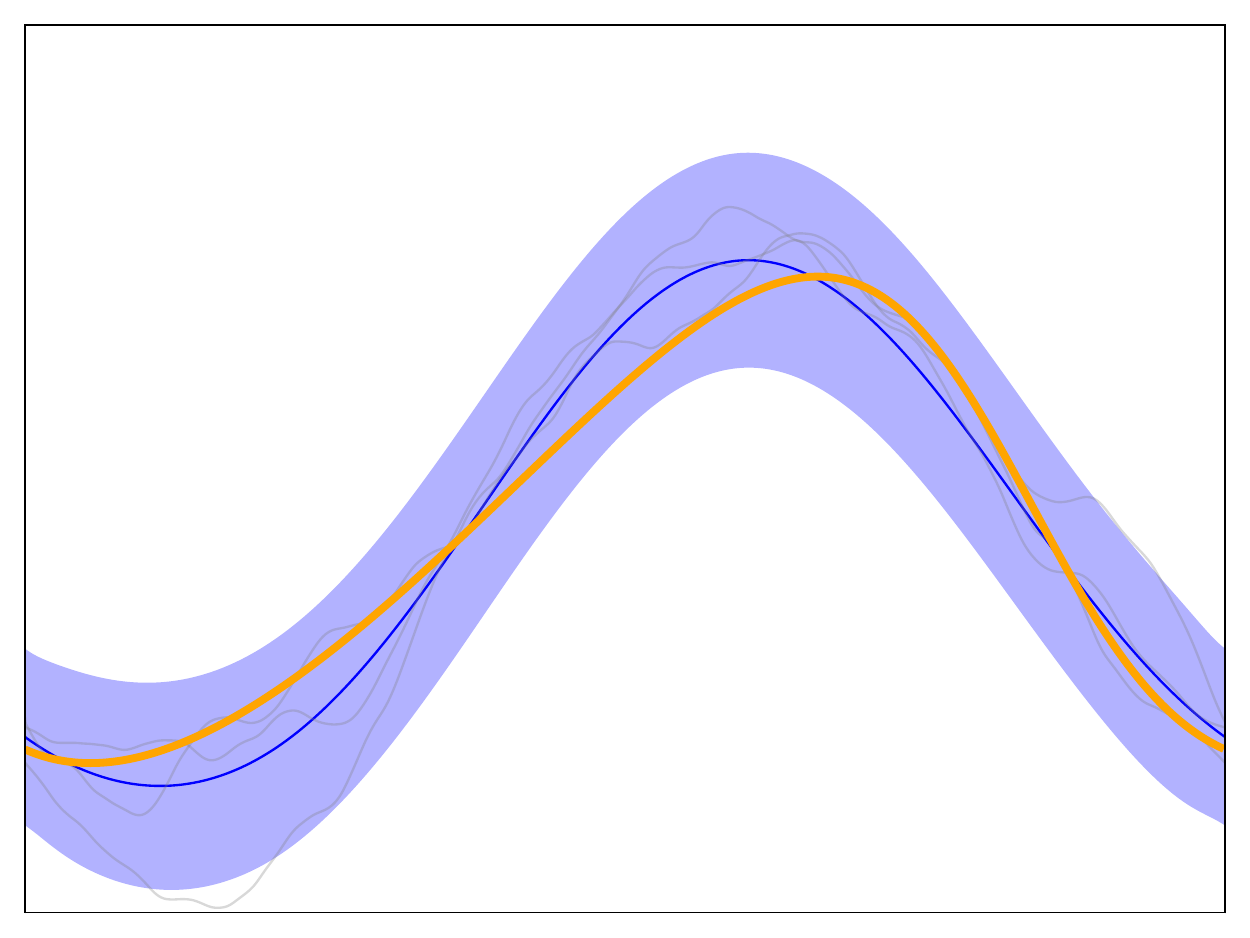}} edge[<-] (advdiff1);

    \node[shape=rectangle, draw=TUdark, minimum width=2cm, rounded corners] (matern1) at (1.5, -2.5) {\includegraphics[width=3.5cm]{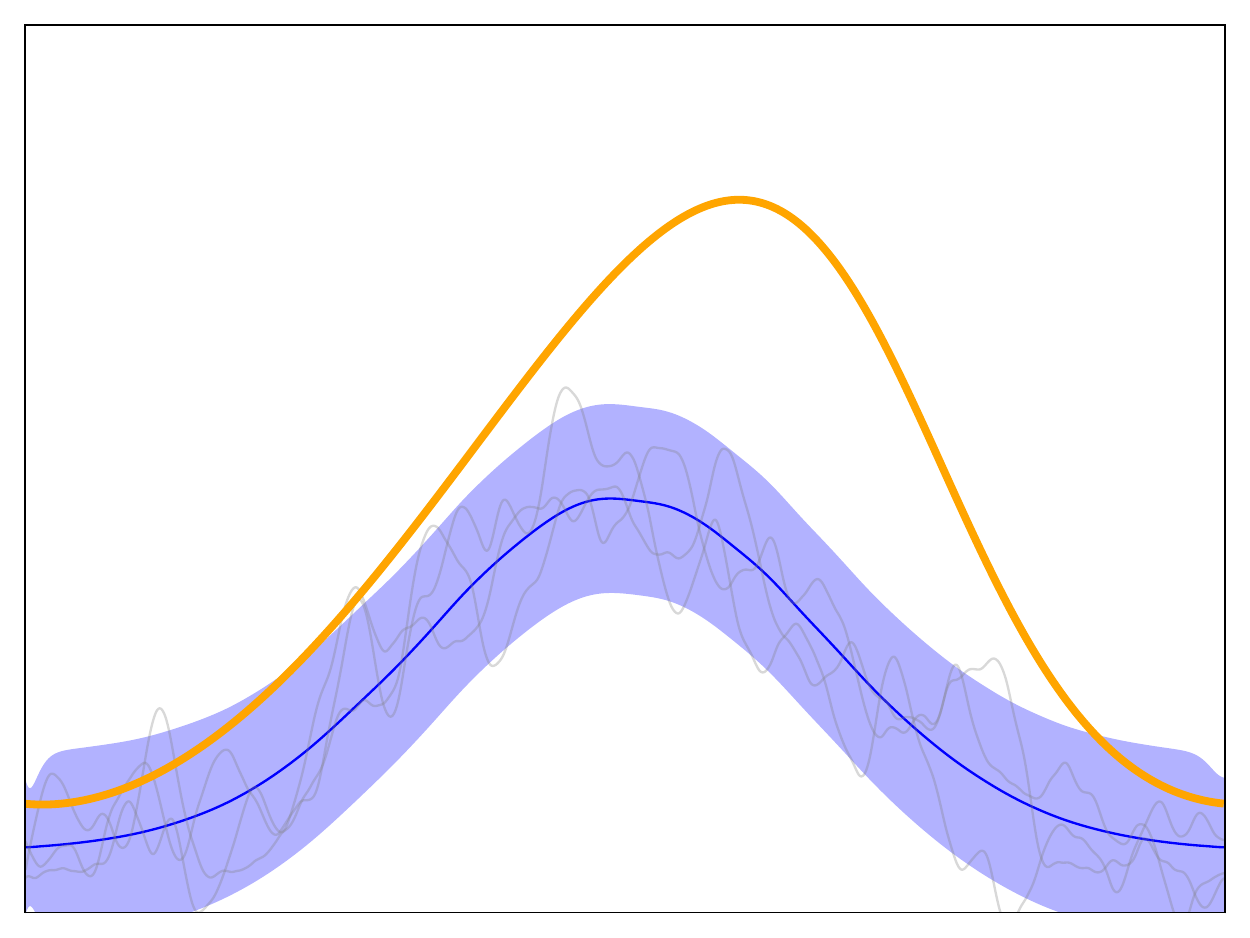}} edge[<-] (prior);
    \node[shape=rectangle, draw=TUdark, minimum width=2cm, rounded corners] (matern2) at (5.5, -2.5) {\includegraphics[width=3.5cm]{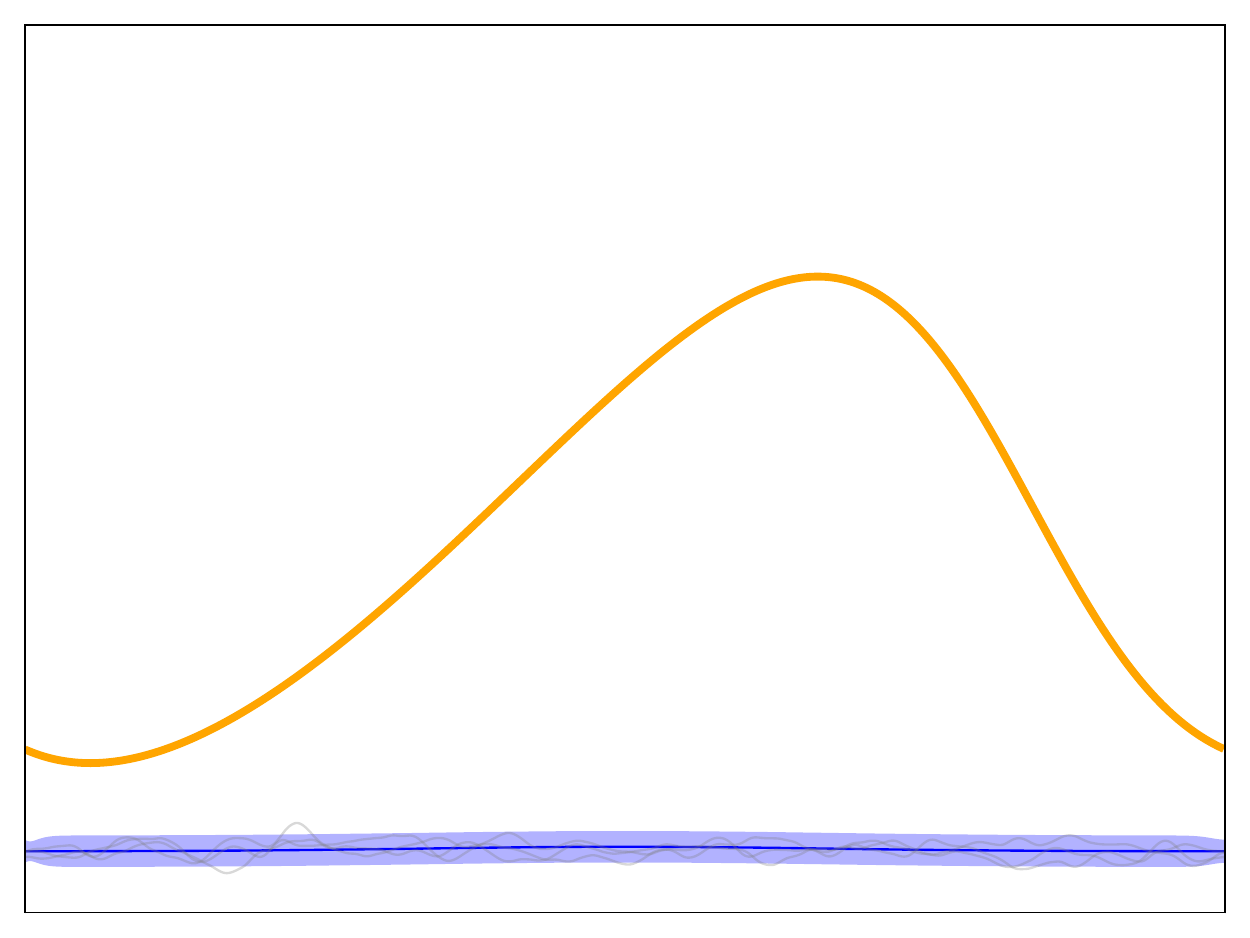}} edge[<-] (matern1);
  \end{tikzpicture}
  \caption{\textbf{Advection-diffusion SPDE priors provide physical information.} Starting from the same initial condition, the top row shows the analytic evolution of the advection-diffusion SPDE prior for Burgers' univariate equation at $t=\SI{1.5}{sec}$ and $t=\SI{3.0}{sec}$.
    For comparison the bottom row shows an analogous stationary product Mat\'ern prior used in prior work.
    Both rows compare to the ground-truth dynamics of a nonlinear Burgers' equation in orange.
}
  \label{fig:prior-comparison}
\end{figure}
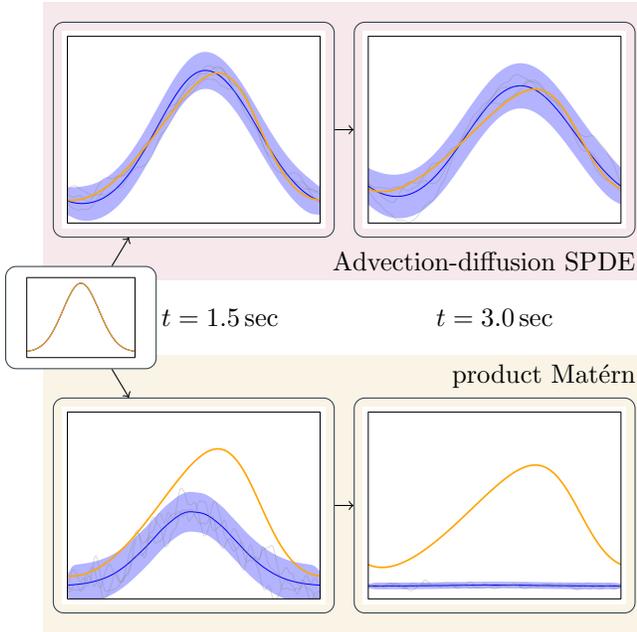

GMRFs are able to approximate Mat\'ern processes, which are a common prior for probabilistic PDE solvers.
While the computational benefits of GMRFs provide a strong motivation for their use (more in \cref{sec:inference-mechanisms}), one may still wonder whether a generic Matérn prior is the optimal choice for a specific probabilistic PDE solver.

\textbf{Priors in terms of SPDE dynamics.}
The Mat\'ern process follows the dynamics of the Whittle-Mat\'ern SPDE.
The dynamics of this SPDE result in posteriors that revert towards the mean as the distance to the data grows, at an effective range and smoothness controllable by $\kappa$ and $\nu$, respectively.
This is a sensible prior for pure regression settings where no mechanistic knowledge about the function to model is available.
However, this is clearly not the case for the application of PDE solvers.
Figure [...] shows the effect of a separable Mat\'ern prior for a Burgers' equation IBVP: The effect of the initial condition simply drops off over time, causing a substantial mismatch to the true dynamics of the problem.

Philosophically, a good prior should include all prior knowledge available about the function of interest.
But this perspective is insufficient for the design of good numerical methods since it does not account for computational cost \citep[\textsection 9.3]{hennig2022probabilistic}.
We know a priori that the function of interest fulfills the PDE, yet encoding this knowledge into the prior incurs computational cost.
In practice, the expressive power of a prior is thus necessarily always limited by computational resources.

\textbf{Beyond the Mat\'ern.} The SPDE formulation introduced above actually allows for a broader perspective than just stationary Matérn processes. At its core, it forms priors as solutions to linear SPDEs.
For the same discretization resolution, a generic Mat\'ern prior is then comparable in computation time to \emph{any} prior based on arbitrary \emph{other}, but still \emph{linear} SPDEs.
We thus propose to construct priors which more closely capture the dynamics of the target (nonlinear) PDE solution through \textbf{linear stochastic proxies}, the probabilistic prior class analogous to linearization of nonlinear PDEs, a concept well-known in the context of numerical PDE solvers.

To exemplify this idea, we return to our running example of Burgers' equation.
Burgers' equation exhibits linear diffusion and \textcolor{red}{nonlinear} advection dynamics:
\begin{align}
\frac{\partial u}{\partial t} + \underbrace{\textcolor{red}{u \frac{\partial u}{\partial x}}}_{\text{Advection}} &= \underbrace{\nu \frac{\partial^2 u}{\partial x^2}}_{\text{Diffusion}}. \label{eq:burgers-pde-adv-diff-explained}
\end{align}
Define the average initial function value $c := \int_{0}^1 u(0, x) dx$ and consider the linear PDE
\begin{equation}
\frac{\partial u}{\partial t} + c \frac{\partial u}{\partial x} = \nu \frac{\partial^2 u}{\partial x^2}. \label{eq:burgers-proxy}
\end{equation}
This is a linear proxy for Burgers' equation that captures the diffusion and the bulk advection.
If we further introduce a stochastic forcing term $\mathcal{W}$ scaled by a parameter $\tau \in \R$, we obtain the advection-diffusion SPDE
\begin{equation}
\frac{\partial u}{\partial t} + c \frac{\partial u}{\partial x} - \nu \frac{\partial^2 u}{\partial x^2} = \tau \mathcal{W}, \label{eq:burgers-proxy-spde}
\end{equation}
which we can use to form a GMRF prior.
\Cref{fig:prior-comparison} shows that this prior indeed captures the general dynamics of the true solution much more effectively.

\textbf{Choosing a linear proxy.}
The success of the above approach relies on our ability to find a linear proxy that approximates the dynamics of the nonlinear PDE of interest well.
Linearization of nonlinear PDEs is well-studied in physics.
For example:
In fluid dynamics, the linearized shallow-water equations find common use \citep{Vallis2017};
Monge-Ampère equations may be solved numerically by iteratively solving (linear) Poisson equations \citep{Benamou2010}, and so on.
Our suggested approach is not limited to any specific linearization technique.
Rather, practitioners may use their domain knowledge to flexibly choose an appropriate proxy for a specific problem.

\textbf{Boundary conditions.}
The FEM discretization of the SPDE involves the specification of boundary conditions.
In spatial statistics, the resulting boundary effects are often undesirable, and a common approach to avoid these effects is to artificially inflate the problem domain \citep{Lindgren2011}.
For a probabilistic PDE solver however, this enables us to embed the boundary condition of the IBVP of interest directly into the prior, as long as the condition is linear.
When assembling the stiffness and mass matrices, we eliminate the corresponding degrees of freedom from the system as in classic FEM.
Having constructed the precision matrix, we may then tune the certainty of the boundary condition through the diagonal entries corresponding to these degrees of freedom.
For more details, refer to \cref{appendix:boundary-conditions}.

\subsection{Spatiotemporal Models}
\label{sec:spatiotemporal-models}
For purely spatial PDEs, the basis function discretization can be chosen analogously to classic FEM solvers.
In spatiotemporal settings however, the GMRF approach differs in the sense that time stepping is not possible and the entire spacetime domain needs to be discretized at once.
Conceptually, time may be treated in the same way as space, and thus a FEM basis may also be used along the time dimension.
In practice, we find that this adds implementation overhead as existing FEM software is not designed for this use case.

\textbf{Implicit Euler discretization.}
Instead, we suggest to use an implicit Euler discretization along time as described by \citet{Clarotto2024} for the special case of an advection-diffusion SPDE.
Following Rothe's method, we first discretize along time through implicit Euler and then discretize the spatial component via FEM.
This leads to a discrete state-space model of the form
\begin{gather}
  \underbrace{(\bm{M} + \Delta t \bm{K})}_{=: \bm{G}} \bm{u}^{(k+1)} = \bm{M} \bm{u}^{(k)} + \sqrt{\Delta t} \bm{M} \bm{\Sigma}^{\nicefrac{1}{2}} \bm{z} \\
  \Leftrightarrow \bm{u}^{(k+1)} = \underbrace{\bm{G}^{-1} \bm{M}}_{=: \bm{A}} \bm{u}^{(k)} + \underbrace{\sqrt{\Delta t} \bm{G}^{-1} \bm{M} \bm{\Sigma}^{\nicefrac{1}{2}}}_{=: \bm{F}^{\nicefrac{1}{2}}} \bm{z}, \label{eq:st-ssm}
\end{gather}

\cref{eq:st-ssm} represents a conditional distribution $\bm{u}^{(k+1)} \mid \bm{u}^{(k)} \sim \mathcal{N}(\bm{A} \bm{u}^{(k)}, \bm{F})$. In combination with an initial distribution $\bm{u}^{(0)} \sim \mathcal{N}(\bm{0}, {\bm{Q}^{(0)}}^{-1})$, we get the joint distribution
\begin{equation}
  \label{eq:spatiotemporal-prior}
  \begin{pmatrix} \bm{u^{(0)}} \\ \vdots \\ \bm{u^{(T)}}\end{pmatrix} \sim \mathcal{N}(\bm{0}, \bm{Q}_{\text{ST}}^{-1}),
\end{equation}
where $\bm{Q}_{\text{ST}}$ is a block tridiagonal precision matrix.
For a detailed derivation, refer to \cref{appendix:spatiotemporal-prior}.

\section{INFERENCE MECHANISMS}
\label{sec:inference-mechanisms}
\cref{sec:prior-construction} described how to construct customized priors suited for probabilistic PDE solvers.
In the following, we describe how to perform Bayesian inference with PDE observations under these priors.

\subsection{Affine conditioning}
\textbf{Information operators.}
In the context of GP-based PDE solvers, \textit{information operators} \citep{Cockayne2019} may be used to condition a GP prior on information about the PDE.
Let $u \sim \mathcal{GP}(\mu, k)$ and $\bm{\mathcal{L}}$ a linear operator acting on the paths of $u$ and taking values in $\R^M$.
Under mild assumptions on $u$ and $\bm{\mathcal{L}}$, the posterior $u \mid \bm{\mathcal{L}}[u] = \bm{b} \sim \mathcal{GP}(\mu_{*}, k_{*})$ is again a GP with known closed forms for $\mu_{*}$ and $k_{*}$ \citep{Pfoertner2022}.
For a PDE $\mathcal{D}[u] = f$ with a linear differential operator $\mathcal{D}$, we can then construct an information operator $\bm{\mathcal{I}}_{\bm{\mathcal{L}}, \bm{b}}[u] := (\bm{\mathcal{L}} \circ \mathcal{D})[u] - \bm{b}$ and obtain a closed-form posterior $u \mid \left(\bm{\mathcal{I}}_{\bm{\mathcal{L}}, \bm{b}}[u] = \bm{0} \right)$.
The most common example is \textit{collocation}: Choose $\bm{x} \in \R^d$ and set $\bm{\mathcal{L}}[u] = u(\bm{x})$ and $\bm{b} = f(\bm{x})$.
This choice ensures that the linear PDE is fulfilled \textit{exactly} at the discrete collocation point $x$.

For general GPs, the closed form for the posterior under such information operators involves applying the linear operator to the covariance function.
In our case, this is simplified through our use of a parametric GP $u(\bm{x}) = \Sigma_{j=1}^N u_j \phi_j(\bm{x})$, as
\begin{align}
  \bm{\mathcal{L}}[u] &= \Sigma_{j=1}^N u_j \bm{\mathcal{L}}[\phi_j] 
  = \bm{L_{\phi}} \bm{u},
  \bm{\mathcal{L}}[u] &= \Sigma_{j=1}^N u_j \bm{\mathcal{L}}[\phi_j] \notag \\
  &= \bm{L_{\phi}} \bm{u},
\end{align}
with $\bm{L}_{\phi} \in \R^{M \times N}$ and $\left(\bm{L}_{\phi}\right)_{ij} = \bm{\mathcal{L}}[\phi_j]_i$, which means that the action of the linear operator on $u$ is induced by its action on the basis functions.

\textbf{Affine conditioning of GMRFs.}
In particular, this means that conditioning a GMRF representation on an information operator is equivalent to conditioning on a transformation under a matrix.
Let $\bm{u} \sim \mathcal{N}(\bm{\mu}, \bm{Q}^{-1})$ be an arbitrary GMRF and
let $\bm{A} \in \R^{M \times N}, \bm{b} \in \R^{M}$.
Under the likelihood $\bm{y} \mid \bm{u} \sim \mathcal{N}(\bm{A} \bm{u} + \bm{b}, \bm{Q_{\epsilon}}^{-1})$, we obtain \citep{Rue2005}:
\begin{align}
  \bm{u} \mid \bm{y} &\sim \mathcal{N}(\bm{\mu}_{u \mid y}, \bm{Q}_{u \mid y}^{-1}), \\
  \bm{Q}_{u \mid y} &= \bm{Q} + \bm{A}^T \bm{Q}_{\epsilon} \bm{A}, \\
\bm{\mu}_{u \mid y} &= \bm{\mu} + \bm{Q}_{u \mid y}^{-1} \bm{A}^T \bm{Q_{\epsilon}} \left( \bm{y} - \left( \bm{A} \bm{\mu} + \bm{b} \right) \right). \label{eq:posterior-mean}
\end{align}
These equations are the key to efficient closed-form conditioning on information operator observations.
Two key examples include
\begin{align}
  \textbf{Collocation: } A_{ij} &:= \mathcal{D}\phi_j(\bm{x}_i) \text{ and } b_i := f(\bm{x}_i)\\
  \textbf{FEM: } A_{ij} &:= \int \langle \phi_i(\bm{x}), \mathcal{D} \phi_j(\bm{x}) \rangle d\bm{x},\\
  \text{and  } b_i &:= \int \langle \phi_i(\bm{x}), f(\bm{x}) \rangle d\bm{x}.
\end{align}
Additionally, we highlight that classic regression also fits into this viewpoint through $A_{ij} := \phi_j(\bm{x_i})$.

\subsection{Moments and sampling}
\label{sec:statistics-and-sampling}
Efficient numerical linear algebra is a crucial component of state-of-the-art PDE solvers.
In the following, we present how GMRFs as a model class naturally enable efficient algorithms for the computation of their key quantities.
Whereas efficiency mostly remains an afterthought for alternative methods for probabilistic PDE solving, GMRF computations are efficient by design.

\textbf{Sparse Cholesky decomposition.}
The precision matrix is sparse, symmetric and positive definite, and thus a sparse Cholesky decomposition $\bm{Q} = \bm{L} \bm{L}^T$ may be computed.
Through efficient node reorderings that minimize fill-in, such decompositions achieve runtime complexities ranging from $\mathcal{O}(N)$ for temporal models to $\mathcal{O}(N^2)$ for spatiotemporal models \citep{Rue2005}.
Once the Cholesky decomposition has been computed, computing the posterior mean in \cref{eq:posterior-mean} requires one sparse forward and one sparse backward solve.
Similarly, if $\bm{z} \sim \mathcal{N}(\bm{0}, \bm{I})$, then $\bm{\mu} + \bm{L}^{-T} \bm{z} \sim \mathcal{N}(\bm{\mu}, \bm{Q}^{-1})$, i.e.\ sampling requires one backward solve.

\textbf{Conjugate gradient method.}
In situations where a sparse Cholesky decomposition is too expensive (due to fill-in), the conjugate gradient (CG) method may be used to solve linear systems iteratively.
This is directly applicable to \cref{eq:posterior-mean}.
To sample using CG, we assume that we have access to a left square root $\bm{L}$ of $\bm{Q}_{u \mid y}$.
Then, if $\bm{z} \sim \mathcal{N}(\bm{0}, \bm{I})$, we get $\bm{L} \bm{z} \sim \mathcal{N}(\bm{0}, \bm{Q}_{u \mid y})$ and thus ${\bm{Q}_{u \mid y}}^{-1} \bm{L} \bm{z} \sim \mathcal{N}(\bm{0}, \bm{Q}_{u \mid y})$, which may be solved using CG.
All models considered in this work have easily accessible left square roots:
For \cref{eq:sparse-prec-matrix} we use $\bm{L} := \bm{K}^{T} \bm{\tilde{M}}^{-\nicefrac{1}{2}}$,
and for the general spatiotemporal prior in \cref{eq:spatiotemporal-prior} we derive a square root in \cref{appendix:spatiotemporal-prior}.
For $\bm{Q} = \bm{A} + \bm{B}$ with $\bm{A} = \bm{L}_{A} \bm{L}_{A}^T$ and $\bm{B} = \bm{L}_{B} \bm{L}_{B}^T$, a left square root is given by $\begin{pmatrix} \bm{L}_A & \bm{L}_B \end{pmatrix}$.

\textbf{Marginal variance computation.}
The marginal variances are given by the diagonal of the covariance matrix, i.e.\ the inverse of the precision matrix.
Marginal variances can be computed without full inversion of the matrix through the Takahashi recursion \citep{RueMarginal2005}.
While the Takahashi recursion is accurate, it also quickly becomes prohibitively expensive.
An alternative is a Rao-Blackwellized Monte Carlo estimator of the variance described by \citet{Siden2018}, which uses the precision matrix to achieve reasonable accuracies even with relatively few samples.

\subsection{Nonlinear PDEs}
With the individual inference step set up, we can now finally treat the general setting of nonlinear PDEs.
Here, we consider a nonlinear function $\bm{f}: \R^{N} \rightarrow \R^{M}$ which induces a likelihood $\bm{y} \mid \bm{u} \sim \mathcal{N}(\bm{f}(\bm{u}), \bm{Q_{\epsilon}}^{-1})$.
For example, $\bm{f}$ may evaluate the PDE residual at $M$ collocation points. 
This yields a posterior density proportional to
\begin{multline}
  \pi(\bm{u}) := \exp \biggl( -\frac{1}{2} (\bm{u} - \bm{\mu} )^T \bm{Q} (\bm{u} - \bm{\mu}) \\ - \frac{1}{2} (\bm{y} - \bm{f}(\bm{u}))^T \bm{Q_{\epsilon}} (\bm{y} - \bm{f}(\bm{u})) \biggr). \label{eq:nonlinear-post-density}
\end{multline}
Then the mode is given by
\begin{equation}
  \bm{\mu_{*}} := \argmax_u \pi(\bm{u}) = \argmin_u \left(- \log \pi(\bm{u}) \right). \label{eq:nonlinear-mode}
\end{equation}
\textbf{Gauss-Newton optimization.}
The right-hand side expression in \cref{eq:nonlinear-mode} can be framed as a nonlinear least-squares problem and solved through Gauss-Newton optimization, which is a common technique also for classic PDE solvers.
Each Gauss-Newton iteration involves a linear system solve with the Gauss-Newton matrix $\bm{H}^{(k)} := \bm{Q} + {\bm{J}^{(k)}}^T \bm{Q_{\epsilon}} \bm{J}^{(k)} \approx - \nabla \nabla^T \log \circ \pi$, where $\bm{J}^{(k)} \in \R^{M \times N}$ is the Jacobian of $\bm{f}$ in iteration $k \in \mathbb{N}$.
Due to the compact basis function representation of $u$, and because $\bm{f}$ is composed of derivative information, the Jacobian is sparse and $\bm{H}^{(k)}$ generally has the same sparsity pattern as $\bm{Q}$. This enables the use of the sparse linear algebra techniques discussed above.

\textbf{Acceleration techniques.} In the sparse Cholesky setting, the node reordering only needs to be computed once and can be reused for all further iterations.
In a CG setting, the Gauss-Newton direction computed in the previous iteration may be reused as an initial guess for CG.
Independently of the linear solver, we employ a backtracking line search to stabilize the optimization.

\textbf{Laplace approximation.}
After converging at a solution for the mode $\bm{\mu_{*}}$, we may then form a Gauss-Newton Laplace approximation of \cref{eq:nonlinear-post-density} as
\begin{equation}
  \bm{x_{*}} \sim \mathcal{N}\left(\bm{\mu_{*}}, \left( \bm{Q} + \bm{J_{*}}^T \bm{Q_{\epsilon}} \bm{J_{*}} \right)^{-1} \right).
\end{equation}
To draw samples and compute variances, we employ the strategies described in \cref{sec:statistics-and-sampling}.

\section{RELATED WORK}
\label{sec:related-work}

\paragraph{GP-based PDE solvers}
There is a growing body of work on GP-based PDE solvers.
These methods can be broadly categorized by the strategy used to discretize the PDE, their applicability to weak or strong form PDEs, and by their ability to solve nonlinear equations.
\citet{Graepel2003LinOpEqGP,Cockayne2017PNPDEInv,Raissi2017LinDEGP} propose collocation-based methods for approximating strong solutions to linear PDEs.
\citet{Owhadi2015BayesianHomogenization} casts numerical homogenization of linear weak-form PDEs as a Bayesian inference problem with a prior Gaussian random field.
\citet{Pfoertner2022} construct a general framework for GP-based solvers for linear PDEs in weak or strong form that is shown to generalize all classical methods of weighted residuals.
\citet{Weiland2024GPFVM} build a GP analogue of the finite volume method.
GP-based methods for nonlinear PDEs using finite-difference discretization \citep{Wang2021BayesNonlinPDE,Kraemer2022PNMOL} and collocation \citep{Chen2021NonlinPDEGP, Chen2024} have also been proposed.

\paragraph{SPDE priors} Our work is certainly not the first to explore alternative SPDE priors.
Particularly in the context of spatiotemporal models, there has been an increasing interest in such models in recent years \citep{Vergara2018, Lindgren2023DEMF, Clarotto2024}.
The work of \citet{Vergara2018} is of particular interest since it also sets a strong focus on physically inspired models.
However, to the best of our knowledge, the relevance and ease of construction of such models in the context of probabilistic PDE solvers has not been explored in prior work.

\paragraph{StatFEM and BFEM}
The statistical finite element method (statFEM) \citep{Girolami2021statFEM} is a framework for solving PDE-constrained inverse problems in a Bayesian fashion.
Just like our work, it employs a FEM-discretized linear SPDE prior.
In contrast to our work, this prior is then not conditioned on an information operator, but instead used to identify the parameters of the SPDE.

The Bayesian finite element method (BFEM) \citep{poot2024bayesian} aims to quantify the discretization error in the finite element method.
It also uses a Gaussian prior defined by a linear SPDE.
However, as opposed to our approach, it then proceeds with the inference procedure in covariance rather than in precision form.

\section{EXPERIMENTS}

\begin{figure*}[t]
  \centering
  \includegraphics[width=1.0\linewidth]{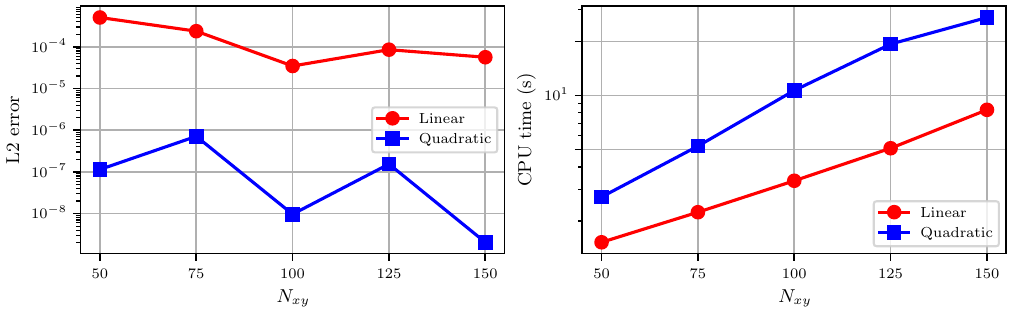}
  \caption{\label{fig:chen-elliptic}\textbf{Performance of our method on the nonlinear elliptic PDE from \citet{Chen2024}}. We evaluate the L2 error and runtime of GMRF-FEM with linear and quadratic basis elements for varying mesh resolutions \(N_{xy}\).}
  \label{fig:chen-elliptic-plot}
\end{figure*}

\label{sec:experiments}
\begin{figure}
  \centering
  \includegraphics[width=\columnwidth]{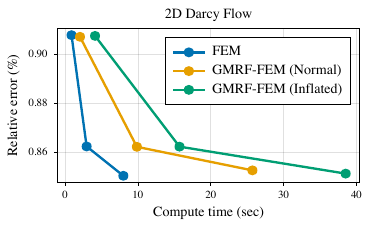}
  \caption{\textbf{GMRF-based methods equal classic FEM solvers in accuracy at reasonable computational overheads.} The plot shows the mean relative error across all problem instances for different discretization resolutions.}
  \label{fig:darcy-work-precision}
\end{figure}

In the following, we evaluate the scaling behavior of GMRF-based PDE solvers as well as the utility of our physics-informed priors.

\textbf{Implementation.}
The code for our experiments is available on GitHub\footnote{\url{https://github.com/timweiland/DiffEqGMRFs.jl}}.
It is based on our own implementation of a framework for GMRFs which utilizes the built-in sparse array routines of Julia\footnote{\url{https://github.com/timweiland/GaussianMarkovRandomFields.jl}}.

\textbf{Datasets.}
We test various aspects of our method on a variety of example problems.
 \citet{LiNeuralOperator2021} introduced a benchmark to evaluate a neural operator.
From this benchmark, we evaluate on 2048 different samples of a \textbf{2D Darcy Flow} problem with Dirichlet boundary conditions, which is a purely spatial PDE with highly discrete variable coefficients, as well as 30 different samples of a \textbf{1D Burgers' equation} IBVP with periodic boundary conditions, which is a common benchmark to test the ability of PDE solvers to handle nonlinearities.
Furthermore, to test the performance of our solver against a state-of-the-art probabilistic solver, we also run experiments on the \textbf{nonlinear elliptic equation} and the \textbf{1D Burgers' equation} presented in \citet{Chen2024}.
\cref{appendix:experimental-details} contains further details on the setup of the experiments.

\subsection{Accuracy and Efficiency}
On the 2D Darcy Flow problem, we start with two versions of a Matérn prior: The first follows the commonly used approach of an inflated boundary (see \cref{sec:physics-informed-priors}) and then manually conditions on the Dirichlet boundary conditions, while the second directly embeds the boundary conditions into the prior as detailed in \cref{sec:physics-informed-priors}.
We then condition both priors on FEM discretizations of Darcy's flow of different resolutions on the interior of the domain.
To validate the hypothesis that our solvers reproduce the accuracy of classic methods and to check the computational overhead of our methods, we also compare to a FEM baseline which uses the same implementation for the assembly of the stiffness matrix.

The work-precision diagram in \cref{fig:darcy-work-precision} demonstrates that GMRF-based solvers with FEM observations indeed reproduce the accuracy of classic FEM solvers, with only minute differences that may be attributed to the noisy observation scheme.
More crucially, \textbf{the runtime of our GMRF-based solver is at the same order of magnitude as FEM}, with a natural overhead arising from added functionality (namely uncertainty quantification).

Furthermore, we observe that there is no difference in accuracy between inflating the boundary and directly embedding the correct boundary conditions into the prior.
However, naturally the inflation of the boundary creates additional degrees of freedom and thus also inflates the size of the precision matrix, increasing the solve time.
For this reason, we find that \textbf{GMRF-based PDE solvers should--if possible--include the boundary conditions in the prior}.

\subsection{Physics-Informed Priors}
\begin{figure}
  \centering
  \includegraphics[width=1.0\linewidth]{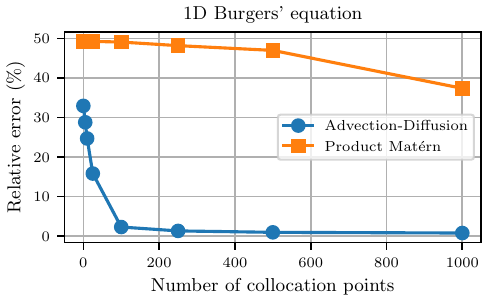}
  \caption{\textbf{Physics-informed priors yield faster convergence in terms of the discretization resolution.} The plot shows the mean relative error across all problem instances for different discretization resolutions for the Burgers' equation IBVPs.}
  \label{fig:burgers-plot}
\end{figure}

We validate the value of physics-informed priors on the Burgers' equation benchmark.
To this end, we start with a separable Matérn prior as well as an advection-diffusion prior, both of which have an embedded periodic boundary.
We then condition on the initial condition and perform Gauss-Newton optimization to solve for nonlinear collocation observations of the PDE.
The results are shown in \cref{fig:burgers-plot}.
We observe that the accuracy for the advection-diffusion prior--which captures the rough dynamics of Burgers' equation-- improves greatly even after only adding a few collocation points.
By contrast, the separable Matérn prior requires many more collocation points to achieve a substantially improved accuracy.
We attribute this to the behavior of the Matérn to revert to the mean, which causes the collocation points to have less of an effect.
These findings validate our intuition that \textbf{physics-informed priors pay off by requiring fewer PDE observations to converge to an accurate posterior}.

\begin{figure*}[t]
  \centering
  \includegraphics[width=1.0\linewidth]{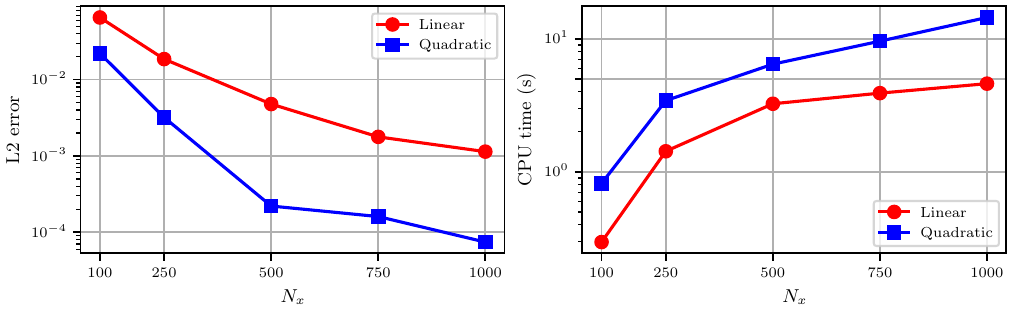}
  \caption{\textbf{Performance of our method on the Burgers' equation example from \citet{Chen2024}}. We evaluate the L2 error and runtime of GMRF-FEM with linear and quadratic basis elements for varying mesh resolutions \(N_x\).}
  \label{fig:chen-burgers-plot}
\end{figure*}

\subsection{Comparison to the state-of-the-art}
In this section, we compare to a state-of-the-art probabilistic PDE solver introduced by \citet{Chen2024}.
Their method is based on finding sparse approximate inverse Cholesky factors to covariance matrices arising from GPs formulated in terms of a covariance function.
By contrast, our method begins with an SPDE perspective, a discretization of which \textit{naturally} yields precision matrices with a controllable sparsity.

\textbf{Nonlinear elliptic equation.}
\citet{Chen2024} present a purely spatial nonlinear elliptic PDE, which we solve as before using FEM observations combined with Gauss-Newton optimization.
\Cref{fig:chen-elliptic-plot} shows that using quadratic elements, our method achieves a high accuracy of 1e-9 L2 error in around 20 seconds.
For comparison: The baseline requires a runtime on the order of hundreds of seconds to achieve an L2 error on the order of 1e-8.
Our method achieves the same order of error in 10 seconds.
This improvement cannot be attributed to hardware differences, but rather to the fundamental design choices of our method, which induces sparsity by design and does not rely on potentially expensive approximate factorizations.

\textbf{1D Burgers' equation.}
Additionally, \citet{Chen2024} evaluate their method on a 1D Burgers' equation.
As before, we use an advection-diffusion prior.
Along space, we use \(N_x \in \mathbb{N}\) finite elements.
We discretize along time using the same time step as the baseline of \(\Delta t = 0.02\).
For the sake of efficiency, we use an implicit Euler discretization to form the prior, and then switch to a Crank-Nicolson discretization for the actual FEM observations of the PDE, \textbf{highlighting the flexibility of our method}.

\Cref{fig:chen-burgers-plot} shows the results.
Our method roughly equals the baseline in terms of accuracy and runtime.
Importantly, \textbf{we want to highlight that our method differs in terms of its treatment of the spatiotemporal dynamics}.
To the best of our knowledge, the baseline method does not propagate uncertainties between time steps.
This enables them to work with \(50\) small kernel matrices, whereas our method works with one sparse block tridiagonal system for the entire spacetime system.
Hence, our method offers a more principled treatment of spacetime that propagates uncertainties along time.

\section{CONCLUSION}
\label{sec:conclusion}
We have presented a framework for efficient probabilistic PDE solvers based on GMRFs.
The SPDE approach enables us to construct physics-informed priors that more closely match the dynamics of the true solution, at comparable computation time compared to classically used stationary models.
Furthermore, GMRFs map inference to sparse linear algebra, allowing for highly efficient computations which enable a comparison to classic solvers, while adding functionality in the form of principled uncertainty quantification.

This necessarily implies an overhead in runtime, which is however massively reduced in contrast to prior work due to the efficient inference mechanisms enabled by the GMRF framework.
Our method matches the high accuracy of classic PDE solvers---which surpasses even state-of-the-art deep learning methods---while providing added functionality in the form of a principled treatment of uncertain data and PDE parameters, all in a machine learning-based method.
We hope that our work can contribute to the development of hybrid methods which mix the high inference speeds of deep learning surrogates with the high accuracy and theoretical guarantees of classic PDE solvers.

\newpage
\subsubsection*{Acknowledgements}
The authors gratefully acknowledge co-funding by the European Union (ERC, ANUBIS, 101123955).
Views and opinions expressed are however those of the author(s) only and do not necessarily reflect those of the European Union
or the European Research Council.
Neither the European Union nor the granting authority can be held responsible for them. PH is supported by the DFG through Project HE 7114/6-1 in SPP2298/2.
PH is a member of the Machine Learning Cluster of Excellence, funded by the Deutsche Forschungsgemeinschaft (DFG, German Research Foundation) under
Germany’s Excellence Strategy – EXC number 2064/1 – Project number 390727645.
The authors also gratefully acknowledge the German Federal Ministry of Education and Research (BMBF) through the Tübingen AI Center (FKZ:01IS18039A); and funds from the Ministry of Science, Research and Arts of the State of Baden-Württemberg.
The authors thank the International Max
Planck Research School for Intelligent Systems (IMPRS-IS) for supporting TW and MP.
The authors are also grateful to Tobias Weber for suggesting improvements to the introduction of this text.

\bibliography{main}

\clearpage
\newpage

\onecolumn
\aistatstitle{Appendix}

\setcounter{section}{0}
\renewcommand{\thesection}{\Alph{section}}

\section{SPATIOTEMPORAL PRIOR}
\label{appendix:spatiotemporal-prior}

\subsection{Construction}
In the following, we describe the spatiotemporal prior construction introduced in \cref{sec:spatiotemporal-models} in more detail.
We follow the construction of \citet{Clarotto2024} for a general spatiotemporal SPDE.

\textbf{Setup.}
Assume a spatiotemporal domain $\mathbb{D} = [0, T] \times \mathbb{D}_s$ and a PDE of the form
\begin{align}
  \frac{\partial u}{\partial t}(t, \bm{x}) + \mathcal{L}_{s}[u](t, \bm{x}) = \mathcal{W}(t, \bm{x}), \rlap{\qquad ($(t, \bm{x}) \in \mathbb{D}$)}
\end{align}
where $\mathcal{L}_s[u]$ is a differential operator that only acts on the spatial dimensions of $u$ and $\mathcal{W}$ is spatiotemporal noise.
More specifically, we assume that the noise has the form
\begin{equation}
  \mathcal{W}(t, \bm{x}) = \mathcal{W}_t(t) \otimes \mathcal{W}_s(\bm{x}),
\end{equation}
where $\mathcal{W}_t$ is Gaussian white noise and $\mathcal{W}_s$ is Gaussian noise.
In practice, to ensure sufficient regularity, we choose $\mathcal{W}_s$ as spatial Matérn noise.

\textbf{Temporal discretization.} Following Rothe's method, we first discretize the temporal dimension through an implicit Euler scheme (which exhibits improved stability properties over explicit Euler schemes).
Consider a temporal domain $[0, T]$ and a partition $0 = t_1 < t_2 < \dots < t_{N_t - 1} < t_{N_t} = T$ with $N_t \in \N$.
Then implicit Euler yields equations of the form
\begin{equation}
  \label{eq:euler-disc-equations}
u^{(i+1)}(\bm{x}) - u^{(i)}(\bm{x}) + (t_{i+1} - t_i) \mathcal{L}_s[u^{(i+1)}](\bm{x}) = \sqrt{t_{i+1} - t_i} \mathcal{W}_s(\bm{x}),
\end{equation}
for $i \in \{1, \dots, N_t - 1\}$.

\textbf{Spatial discretization.}
Next, we assume an appropriate FEM basis function representation for the spatial dimensions.
In practice, we use Lagrange polynomials of sufficient order such that forming observations using collocation or FEM for the PDE of interest is possible subsequently.
We write $u^{(i)} = \sum_{j=1}^N u^{(i)}_j \phi_j$ and integrate \cref{eq:euler-disc-equations} against test functions, which yields
\begin{align}
  &\bm{M} \bm{u}^{(i+1)} - \bm{M} \bm{u}^{(i)} + (t_{i+1} - t_i) \bm{K} \bm{u}^{(i+1)} = \sqrt{t_{i+1} - t_i} \bm{M} \bm{\Sigma^{\nicefrac{1}{2}}} \bm{z} \\
  \iff &\underbrace{\left( \bm{M} + (t_{i+1} - t_i) \bm{K} \right)}_{=: \bm{G_i}} \bm{u}^{(i+1)} = \bm{M} \bm{u}^{(i)} + \underbrace{\sqrt{t_{i+1} - t_i} \bm{M} \bm{\Sigma^{\nicefrac{1}{2}}}}_{=: \bm{F_i}^{\nicefrac{1}{2}}} \bm{z} \\
  \iff & \bm{u}^{(i+1)} = \underbrace{\bm{G_i}^{-1} \bm{M}}_{=: \bm{A_i}} \bm{u}^{(i)} + \bm{F_i}^{\nicefrac{1}{2}} \bm{z}
\end{align}
where $\bm{M}$ is the mass matrix, $\bm{K}$ is the stiffness matrix corresponding to $\bm{L}_s$, $\bm{\Sigma}^{\nicefrac{1}{2}}$ is a square root of the covariance matrix resulting from a FEM discretization of the spatial Matern noise process, and $\bm{z} \sim \mathcal{N}(\bm{0}, \bm{I})$.

\textbf{State-space model.}
Assume an initial distribution with precision $\bm{Q_1}$ and mean zero.
In practice, we use a GMRF approximation to a spatial Matérn process.
We obtain the state-space model
\begin{align}
  \bm{u}^{(1)} &\sim \mathcal{N}(\bm{0}, \bm{Q_1}), \\
  \bm{u}^{(i+1)} \mid \bm{u}^{(i)} &\sim \mathcal{N}(\bm{A_i} \bm{u}^{(i)}, \bm{F_i}).
\end{align}
This yields the joint distribution
\begin{equation}
  \begin{pmatrix} \bm{u^{(1)}} \\ \vdots \\ \bm{u^{(N_t)}}\end{pmatrix} \sim \mathcal{N}(\bm{0}, \bm{Q}_{\text{ST}}^{-1}),
\end{equation}
with block tridiagonal precision matrix
\begin{equation}
  \label{eq:block-tridiag-prec}
  \bm{Q_{\text{ST}}} :=
\begin{pmatrix}
\bm{Q_1}^{-1} + \bm{A_1}^T \bm{F_1}^{-1} \bm{A_1} & -\bm{A_1}^T \bm{F_1}^{-1} & \bm{0} & \dots & \bm{0} \\
-\bm{F_1}^{-1} \bm{A_1} & \bm{F_1}^{-1} + \bm{A_2}^T \bm{F_2}^{-1} \bm{A_2} & -\bm{A_2}^T \bm{F_2}^{-1} & \ddots & \vdots \\
\bm{0} & \ddots & \ddots & -\bm{A_{N_t - 2}}^T \bm{F_{N_t - 2}}^{-1} & \bm{0} \\
\vdots & \ddots & -\bm{F_{N_t - 2}}^{-1} \bm{A_{N_t - 2}} & \bm{F_{N_t - 2}}^{-1} + \bm{A_{N_t - 1}}^T \bm{F_{N_t - 1}}^{-1} \bm{A_{N_t - 1}} & -\bm{A_{N_t - 1}}^T \bm{F_{N_t - 1}}^{-1} \\
\bm{0} & \dots & \bm{0} & -\bm{F_{N_t - 1}}^{-1} \bm{A_{N_t - 1}} & \bm{F_{N_t - 1}}^{-1}
\end{pmatrix}.
\end{equation}
For a temporal discretization with uniform step size, we have $\bm{F_1} = \dots = \bm{F_{N_t - 1}}$ and $\bm{A_1} = \dots = \bm{A_{N_t - 1}}$.

\subsection{Square root}
For efficient sampling when using the conjugate gradient method, a left square root of the precision matrix is helpful.
Given square roots of $\bm{Q_1}$ and $\bm{F_i}^{-1}$ ($i \in \{1, \dots, N_t - 1\}$) -- which are available in closed form by construction -- we can construct a block bidiagonal square root of \cref{eq:block-tridiag-prec} as
\begin{equation}
  \bm{Q_{\text{ST}}}^{\nicefrac{1}{2}} :=
  \begin{pmatrix}
    \bm{Q_1}^{-\nicefrac{1}{2}} \mid \bm{A_1}^T \bm{F_1}^{-\nicefrac{1}{2}} & \bm{0} & \dots & \dots & \bm{0} \\
    \bm{0} \mid -\bm{F_1}^{-\nicefrac{1}{2}} & \bm{0} \mid \bm{A_2}^T \bm{F_2}^{-\nicefrac{1}{2}} & \bm{0} & \dots & \vdots \\
    \bm{0}  & \ddots & \ddots &  \ddots & \vdots \\
    \bm{\vdots}  & \ddots & \ddots &  \bm{0} \mid \bm{A_{N_t - 1}}^T \bm{F_{N_t - 1}}^{-\nicefrac{1}{2}} & \bm{0} \\
    \bm{0}  & \dots & \bm{0} &  \bm{0} \mid -\bm{F_{N_t - 1}}^{-\nicefrac{1}{2}} & \bm{0} \\
  \end{pmatrix},
\end{equation}
which utilizes wide blocks to treat the sums on the diagonal of \cref{eq:block-tridiag-prec}.

\newpage

\section{BOUNDARY CONDITIONS}
\label{appendix:boundary-conditions}

\textbf{Setup.}
In the following, we describe how to implement GMRFs that fulfill linear constraints of the form
\begin{equation}
  \label{eq:linear-constraint}
  u_{k} = \sum_{i=1}^{N_k} c_i u_{h(i)},
\end{equation}
where $k \in \{1, \dots, N\}$ is the index of the degree of freedom that is constrained, $N_k < N$, $\bm{c} \in \R^{N_k}$ and $h: \{1, \dots, N_k\} \rightarrow \{1, \dots, N\}$ maps to the indices of the degrees of freedom that $u_k$ depends on.
Note that this setup includes
\begin{align}
  \textbf{Homogeneous Dirichlet boundary: } &N_k = 0, \\
  \textbf{Periodic boundary: } &N_k = 1, c_1 = 1, h(1) = \textit{mirror}(k),
\end{align}
where the function $\textit{mirror}$ maps degrees of freedom to their appropriate counterparts on the boundary of the domain.

\textbf{Adding constraints to linear systems.}
Any linear system involving the degrees of freedom needs to be adapted to include the linear constraints.
Consider a linear system of the form
\begin{equation}
  \bm{A} \bm{u} = \bm{b}.
\end{equation}
The $i$-th row of this system is the linear equation
\begin{equation}
  \sum A_{ij} u_j = b_i.
\end{equation}
Inserting \cref{eq:linear-constraint} yields
\begin{align}
  &\sum_{l=1}^{N_k} A_{ik} c_l u_{h(l)} + \sum_{j \neq k} A_{ij} u_j = b_i \\
  \iff & \sum_{l=1}^{N_k} (A_{ik} c_l + A_{i,h(l)}) u_{h(l)} + \sum_{j \neq \{k\} \cup h(\{1, \dots, N_k\})} A_{ij} u_j = b_i.
\end{align}
In other words, the linear constraint may be enforced by adding scaled copies of column $k$ to the columns indexed by $h(\{1, \dots, N_k\})$, and then setting column $k$ to zero.

\textbf{FEM matrix assembly.}
The constraints also need to be included in the matrix assembly process.
Here, we first follow the process described above for a general linear system, and then we set $A_{kk} = 1$, $A_{kj} = 0$ ($j \neq k$).
Recall that the construction of a GMRF stems from the relation
\begin{equation}
  \bm{K} \bm{u} \sim \mathcal{N}(\bm{0}, \bm{M}).
\end{equation}
We apply the constraints to both $\bm{K}$ and $\bm{M}$.
If we now further set $M_{kk} = \varepsilon_k^2$, then we obtain $u_k \sim \mathcal{N}(0, \varepsilon_k^2)$.
$\varepsilon_k^2$ may be used to model a noisy boundary condition:
Consider the transformed random variable $\bm{\tilde{u}}$ with $\tilde{u}_j = u_j$ ($j \neq k$) and $\tilde{u}_k = u_k + \sum_{i=1}^{N_k} c_i u_{h(i)}$. Then $\tilde{u}_k \mid (\tilde{u}_{h(1)}, \dots, \tilde{u}_{h(N_k)}) \sim \mathcal{N}(\Sigma_{i=1}^{N_k} c_i \tilde{u}_{h(i)}, \varepsilon_k^2)$. In the limit of $\varepsilon_k^2 \to 0$ this models a hard, deterministic boundary condition.

\newpage

\section{EXPERIMENT DETAILS}
\label{appendix:experimental-details}
The Darcy Flow experiment was run on an internal compute cluster on 8 cores of an Intel Xeon Gold 6240 CPU.
All other experiments were run on a 2021 M1 Max MacBook
Pro.

\subsection{Darcy Flow}
\begin{table*}[t]
  \small
  \centering
\begin{tabular}{l S[separate-uncertainty=true] S[separate-uncertainty=true] S[separate-uncertainty=true] S[separate-uncertainty=true]}
  \toprule
                    & {Compute time (sec)} & {Relative error (\%)} & {Time per sample (sec)} & {Time for std. dev. (sec)} \\
  \midrule
GMRF-FEM (Normal)   & 25.71              & 0.85 \pm 0.05        & 0.51 \pm 0.17          & 33.89 \pm 4.08             \\
GMRF-FEM (Inflated) & 38.51              & 0.85 \pm 0.05        & 0.65 \pm 0.19          & 44.73 \pm 4.67            \\
FEM                 & 8.00               & 0.85 \pm 0.05        & {-}                     & {-}                        \\
FNO                 & {-}                  & 0.98                & {-}                     & {-}                        \\
\bottomrule
\end{tabular}
\caption{Detailed benchmark results for 2D Darcy Flow.}
\label{table:darcy-detailed-results}
\end{table*}

\textbf{Dataset.}
We use the Darcy flow dataset introduced in \citet{LiNeuralOperator2021}, which consists of 2048 samples in total.

\textbf{FEM discretization.}
We use a structured mesh of the unit square consisting of quadratic triangular elements with Lagrange basis functions. We regulate the mesh resolution by specifying the number of elements along each dimension; in practice, we use the same number of elements for both spatial dimensions.

\textbf{FEM baseline.}
For the FEM baseline, we ensure a fair comparison by using the same stiffness matrix assembly code as for GMRF-FEM, which is based on Ferrite.jl \citep{Carlsson2024Ferrite}.
Afterwards, we compute a sparse Cholesky decomposition since the stiffness matrix is symmetric and positive definite for this problem.
Finally, we use the Cholesky decomposition to solve for the degrees of freedom.

\textbf{GMRF-FEM prior.}
For the prior of GMRF-FEM, we use a GMRF approximation of a two-dimensional Matérn covariance with an effective range of $N_{xy}^{-\nicefrac{1}{2}}$, where $N_{xy}$ is the number of elements per dimension, and with $\nu = 2$. In practice, we find that the accuracy of the solution is not very sensitive to the prior hyperparameters due to the strong weight of the observations in this setup.

\textbf{Boundary inflation.}
For the version of GMRF-FEM with an inflated boundary, we inflate the boundary by a width of 0.15 and increase the size of the external elements to up to twice the size of the internal elements.

\begin{table*}[t]
  \small
  \centering
\begin{tabular}{lSSS}
  \toprule
                    & {$N_{xy} = 100$} & {$N_{xy} = 200$} & {$N_{xy} = 300$}  \\
  \midrule
GMRF-FEM (Normal)   & 4.79              & 10.37        & 24.71                       \\
GMRF-FEM (Inflated) & 6.56             & 14.61        & 35.87 \\
\bottomrule
\end{tabular}
\caption{Prior construction times for Darcy flow (in seconds).}
\label{table:darcy-prior-construction-times}
\end{table*}

\textbf{Compute time.}
The compute times in \cref{fig:darcy-work-precision} include the time required to assemble the stiffness matrix for both methods, as well as the linear solve for FEM and the time required to condition the prior on the FEM observations for GMRF-FEM.
Note that we do not include the time required to form the prior for GMRF-FEM, which we only form once at the start to save resources.
\cref{table:darcy-detailed-results} contains more detailed results, and \cref{table:darcy-prior-construction-times} contains the concrete times for the prior construction.

\subsection{Burgers' equation}
\textbf{Dataset.}
We use the Burgers' equation dataset introduced in \citet{LiNeuralOperator2021} and subsample it down to 30 samples.

\textbf{FEM discretization.}
We use a structured mesh of the unit interval consisting of quadratic line elements of equal size with Lagrange basis functions.
For the implicit Euler discretization, we use exactly the time points used for evaluation of the solution, which uses a uniformly spaced partition of the temporal domain $[0.0, 1.0]$ with 101 time points.
In all cases, our prior is based on a spatial discretization consisting of 750 elements.

\textbf{Product Matérn prior.}
For the Product Matérn prior, we use $\nu = \nicefrac{7}{2}$ for the spatial part and $\nu = \nicefrac{1}{2}$ for the temporal part.
We use a temporal range of $3.0$ and a spatial range of $0.02$.

\textbf{Advection-Diffusion prior.}
For the physics-informed prior, we use the model of \cref{eq:burgers-proxy-spde}, which is a special case of the model described in \citet{Clarotto2024}.
For the spatial noise we use a Matérn process with $\nu = \nicefrac{3}{2}$ and a range of $0.02$.

\begin{table*}[t]
  \small
  \centering
\begin{tabular}{S S[separate-uncertainty=true] S[separate-uncertainty=true]}
  \toprule
  {Number of collocation points} & {Advection-Diffusion} & {Product Matern}  \\
  \midrule
0 & 32.94 \pm 10.43 & 49.25 \pm 7.49 \\
5 & 28.79 \pm 8.61 & 49.25 \pm 7.5 \\
10 & 24.7 \pm 7.08 & 49.24 \pm 7.49 \\
25 & 15.83 \pm 5.01 & 49.23 \pm 7.49 \\
100 & 2.32 \pm 2.05 & 49.06 \pm 7.47 \\
250 & 1.34 \pm 1.34 & 48.16 \pm 7.33 \\
500 & 0.98 \pm 1.0 & 46.95 \pm 7.04 \\
1000 & 0.8 \pm 0.54 & 37.37 \pm 4.5 \\
  \bottomrule
\end{tabular}
\caption{Mean and standard deviations of the relative errors of the different methods applied to Burgers' equation, in \%.}
\label{table:burgers-prior-comparison}
\end{table*}

\textbf{Quantitative results}. \cref{table:burgers-prior-comparison} shows the concrete values used in \cref{fig:burgers-plot}.

\subsection{Comparison to state-of-the-art probabilistic solver}
\textbf{Nonlinear elliptic PDE.}
We consider the nonlinear elliptic PDE as presented in \citet{Chen2024}:
\begin{equation}
\begin{cases}
-\Delta u + u^3 = f & \text{in } \Omega := [0, 1]^2, \\
u = g & \text{on } \partial\Omega,
\end{cases}
\end{equation}
with the manufactured solution
\[
u(\mathbf{x}) = \sum_{k=1}^{600} \frac{1}{k^6} \sin(k \pi x_1) \sin(k \pi x_2).
\]

For \(N_{xy} \in \mathbb{N}\), we employ our method with \(N_{xy}^2\) finite elements.
We start with a Mat``ern prior with effective range \(0.1\) and \(\nu = 2\).
Then we condition on the boundary points with a uniform spacing of \(0.001\) and perform Gauss-Newton optimization on FEM observations of the PDE until either the Newton decrement subceeds 1e-5 or a maximum number of \(10\) steps is reached.

\textbf{1D Burgers' equation.}
We consider the nonlinear Burgers' equation problem presented in \citet{Chen2024}:
\begin{gather}
\partial_t u + u \partial_x u - 0.001 \partial^2_x u = 0, \quad \forall(x, t) \in (-1, 1) \times (0, 1), \\
u(x, 0) = -\sin(\pi x), \\
u(-1, t) = u(1, t) = 0.
\end{gather}
The ground-truth solution is computed by a Cole-Hopf transformation via the same code used by \citet{Chen2024}.

We run the Gauss-Newton optimization until either the Newton decrement subceeds 1e-5 or a maximum number of \(30\) steps is reached.

\end{document}